\documentclass[conference,a4paper]{IEEEtran}

\usepackage{times}
\usepackage{epsfig}
\usepackage{graphicx}
\usepackage{amsmath}
\usepackage{amssymb}
\usepackage{booktabs}
\usepackage{multirow}
\usepackage{array}
\usepackage{url}
\usepackage[table]{xcolor}
\usepackage{caption}
\usepackage{placeins}
\usepackage{CJKutf8}
\definecolor{hlpink}{rgb}{1.0,0.75,0.80}

\newcommand{\fg}{\mathrm{fg}}
\newcommand{\bg}{\mathrm{bg}}

\newcommand{\precision}{\mathrm{precision}}

\newcommand{\bestval}[1]{\textcolor{black}{{\usefont{OT1}{cmr}{bx}{n}#1}}}

\title{\Large OT-BLS: Optimal-Transport-based Backbone Layer Selection\\ for Frozen DINOv3 in Cross-Domain Few-Shot Segmentation}

\author{
\IEEEauthorblockN{Natsuo Okamoto, Kazuki Kobayashi, Yoshio Kato, Shuhei Tarashima}
\IEEEauthorblockA{NTT DOCOMO BUSINESS, Inc., Tokyo, Japan \\
natsuo.okamoto@ntt.com, kzk.kobayashi@ntt.com, yoshio.kato@ntt.com, tarashima@acm.org}
}

\begin{document}
\maketitle

\begin{abstract}
Cross-domain few-shot segmentation (CD-FSS) aims to segment a target class in images from a previously unseen domain using only a few annotated support examples. 
Training-free CD-FSS methods, which typically leverage vision foundation models such as DINOv3,
achieve competitive performance without parameter updates.
However, the optimal DINOv3 layer for mask prediction varies across support--query episodes.
Although several layer-selection criteria have been proposed, they do not explicitly evaluate whether the predicted query foreground and background labels agree with the corresponding support labels.
To address this limitation, we propose a novel layer-selection criterion, Optimal-Transport-Based Backbone Layer Selection (OT-BLS).
Our key insight is that an effective layer-selection criterion can be derived from the optimal-transport correspondence between query and support patch-token features. 
Based on this insight, we first estimate the transport plan using a well-established optimal transport algorithm and then compute a layer-selection score by aggregating the transported mass according to the foreground--background partition defined by each candidate query mask.
The resulting score reflects how well the predicted foreground and background of the query correspond to those of the support.
Experiments on eight benchmark datasets demonstrate the effectiveness of OT-BLS. It outperforms the conventional last-layer baseline in 12 of 16 dataset/shot settings. Furthermore, OT-BLS improves the average mIoU from 55.72 to 57.62 and achieves a 2.99-point improvement over the best-performing existing layer-selection criterion while running 3.7--14.6$\times$ faster.
\end{abstract}

\vspace{-0.2em}

\section{Introduction}
Cross-domain few-shot segmentation (CD-FSS) aims to segment a target class in query images given only a few annotated support examples of that class.
The support and query images come from a target domain that differs from the domain used to train the model.
It is especially useful in domains such as medical, underwater, and satellite imagery, where dense pixel-level annotation is costly.
CD-FSS is more challenging than ordinary few-shot segmentation (FSS) because domain differences make support--query correspondence ambiguous~\cite{ma2025hera}.

In addition to CD-FSS methods that assume additional training or domain adaptation~\cite{nie2024cross,herzog2024adapt,liu2023matcher,ma2025hera}, training-free methods based on pre-trained vision foundation models have recently attracted attention.
In this setting, the backbone is kept frozen, and no parameters are fine-tuned for the target domain or for each support--query episode.
Because no parameter updates are required, these methods avoid the per-domain optimization and model-management costs incurred by fine-tuning.
Keeping the backbone frozen also preserves the generalization of the pre-trained features, which fine-tuning on a few target examples can weaken through overfitting.
Zakir \textit{et al.}~\cite{zakir2026revealingsemanticselectiongap} reported that frozen DINOv3~\cite{simeoni2025dinov3} can be competitive with CD-FSS methods that require additional training.
Interestingly, they also showed that the last layer is not always the best choice for query mask prediction.
In many episodes, an intermediate layer yields a more accurate mask.

Existing criteria~\cite{zakir2026revealingsemanticselectiongap} score candidate layers using indirect proxies such as prediction confidence, feature separability, mask reconstruction consistency, or feature-statistic similarity.
These proxies do not directly measure whether the foreground predicted on the query agrees with the support foreground and is clearly separated from the background.
This limitation can cause a criterion to select a layer that performs worse than the simple last-layer baseline, which always uses the last-layer features.

This observation motivates a new layer-selection criterion for CD-FSS.
We propose Optimal-Transport-Based Backbone Layer Selection (OT-BLS).
OT-BLS scores how well the predicted query foreground and background correspond to the matching support foreground and background, and how clearly the predicted foreground is separated from the background.
To compute this score, OT-BLS measures how much query patch-token mass is transported to corresponding support regions under the candidate mask for each layer.
Specifically, OT-BLS first estimates a transport plan between query and support patch-token features using a well-established optimal transport (OT) algorithm, and then computes the score by aggregating the transported mass according to the foreground--background partition defined by each candidate query mask.
Experiments on eight benchmarks demonstrate that OT-BLS outperforms the last-layer baseline in 12 of 16 dataset/shot settings, improves the average mIoU from 55.72 to 57.62, and runs faster than the best-performing existing criterion.

\section{Optimal-Transport-Based Backbone Layer Selection (OT-BLS)}
\label{sec:otbls}
Given a query image and support images, we use the predictor of Zakir \textit{et al.}~\cite{zakir2026revealingsemanticselectiongap} to generate one candidate mask for each DINOv3 layer.
Our goal is to select the best mask from these candidates.
The predictor aligns support masks to the DINOv3 patch grid, computes cosine-similarity maps from support-derived prototypes, and binarizes these maps to obtain $\hat{m}_q^{(l)}$.
OT-BLS only selects among these candidates and does not alter the predictor.
Following existing layer-selection methods~\cite{zakir2026revealingsemanticselectiongap}, layer selection can be written as selecting the candidate layer with the largest scalar score $s^{(l)}$ from the search set $\mathcal{L}$,
\begin{equation}
l^* = \mathop{\arg\max}_{l\in\mathcal{L}} s^{(l)}.
\label{eq:selected_layer}
\end{equation}
OT-BLS differs from prior criteria in how this score is defined.
Let $x_q$ denote a query image and let $\mathcal{S}=\{(x_{s,k},m_{s,k})\}_{k=1}^{K}$ denote the $K$ support examples.
Each ground-truth mask $m_{s,k}\in\{0,1\}^{H\times W}$ is a binary pixel-level annotation of the target class.
We treat pixels of the target class as foreground and all other pixels as background.

Layer $l$ denotes the output of the $l$-th transformer block, and $L$ denotes the last-layer index.
When we form token-level index sets below, pixel masks and candidate masks are resized to the patch-token grid by nearest-neighbor interpolation.

\begin{figure}[t]
\centering
\includegraphics[width=0.8\linewidth]{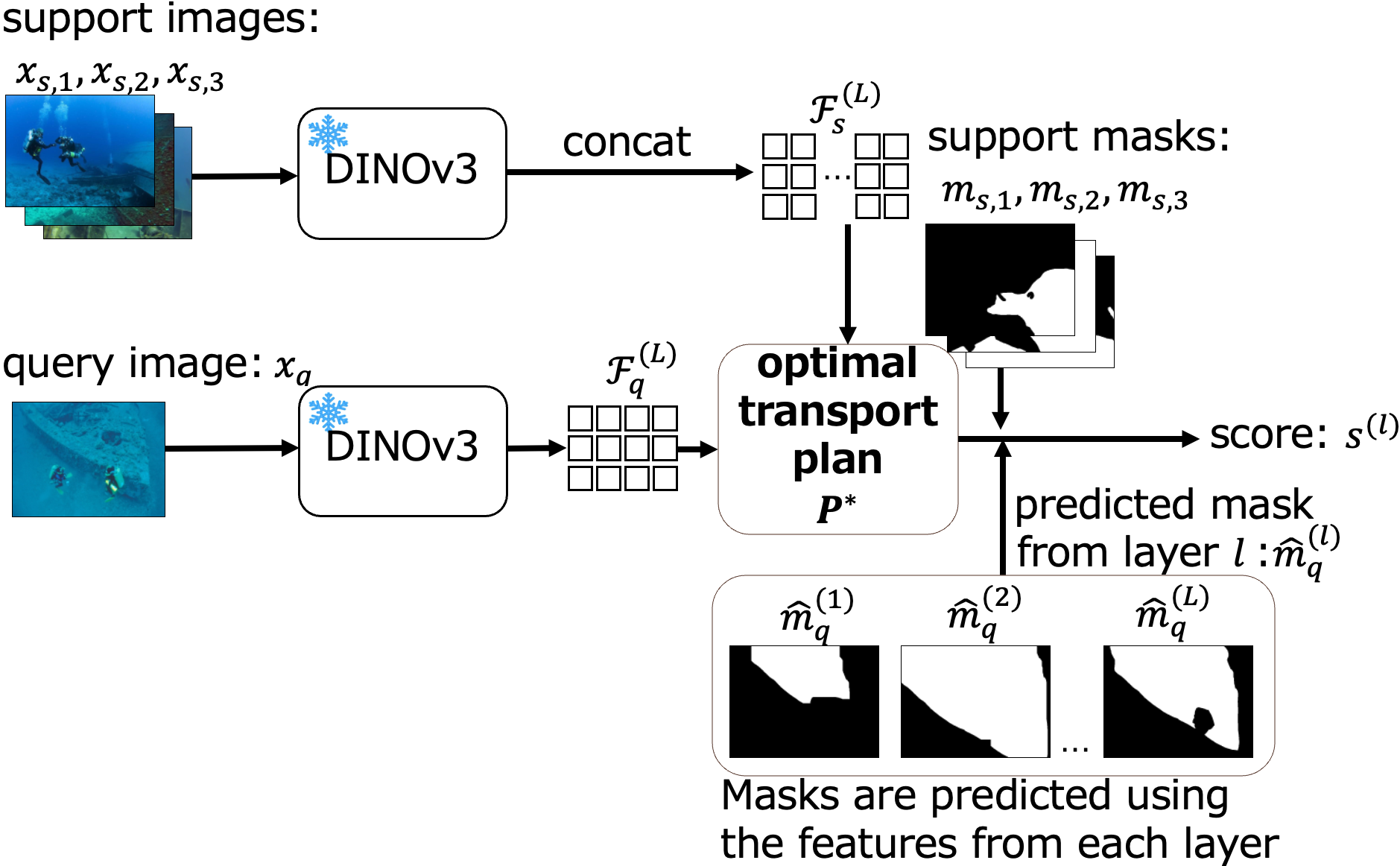}
\caption{Overview of OT-BLS for the case $K=3$.}
\label{fig:proposal_method}
\vspace{-0.7em}
\end{figure}

Figure~\ref{fig:proposal_method} gives an overview of OT-BLS.
We first describe the OT-BLS selection criterion and then explain how to compute it from the optimal transport plan.

\begin{table*}[!t]
\centering
\scriptsize
\setlength{\tabcolsep}{3pt}
\caption{1-shot and 5-shot performance of each method (mIoU, \%).
The best value among all methods except Oracle, including the last-layer baseline, is shaded light pink.
The best value among the selection strategies, excluding Oracle and the last-layer baseline, is bold.
Avg. denotes the mean mIoU over all 16 settings.
Rank denotes the corresponding average rank over the same selection strategies.
COCO-20i values are averaged over four folds.
Oracle and the last-layer baseline are excluded from the rank computation.}
\label{tab:main}
\resizebox{\textwidth}{!}{
\begin{tabular}{cl*{8}{cc}cc}
\hline
& & \multicolumn{2}{c}{DeepGlobe}
& \multicolumn{2}{c}{ISIC}
& \multicolumn{2}{c}{SUIM}
& \multicolumn{2}{c}{Chest X-ray}
& \multicolumn{2}{c}{FSS-1000}
& \multicolumn{2}{c}{SpaceNetv2}
& \multicolumn{2}{c}{Kvasir-SEG}
& \multicolumn{2}{c}{COCO-20i}
& \multicolumn{2}{c}{Overall} \\
\cline{3-20}
& Selection strategy
& 1-shot & 5-shot
& 1-shot & 5-shot
& 1-shot & 5-shot
& 1-shot & 5-shot
& 1-shot & 5-shot
& 1-shot & 5-shot
& 1-shot & 5-shot
& 1-shot & 5-shot
& Avg. & Rank \\
\hline
& Oracle (GT-guided upper bound)
& 57.55 & 67.02
& 67.19 & 74.34
& 61.60 & 69.41
& 76.57 & 76.72
& 86.79 & 88.14
& 39.59 & 42.78
& 54.99 & 61.87
& 54.55 & 62.92
& 65.13 & -- \\
\hline
& Last-layer baseline
& \cellcolor{hlpink}49.02 & \cellcolor{hlpink}59.68
& \cellcolor{hlpink}52.19 & 61.61
& 50.68 & 61.12
& 60.65 & 61.25
& 82.51 & 84.68
& 31.65 & 34.23
& 43.53 & 54.13
& 46.16 & 58.49
& 55.72 & -- \\
\hline
\multirow{8}{*}{\rotatebox{90}{Layer selection}}
& Entropy
& 38.33 & 44.10
& 35.68 & 38.59
& 25.57 & 30.24
& 32.45 & 29.74
& 37.50 & 42.92
& 29.39 & 32.99
& 22.05 & 23.39
& 11.70 & 14.43
& 30.57 & 7.63 \\
& Regularized Patch-Norm Ratio
& 36.84 & 47.06
& 42.32 & 50.43
& 29.91 & 36.02
& 49.31 & 51.37
& 44.61 & 52.84
& 26.86 & 27.81
& 23.99 & 23.51
& 13.53 & 18.21
& 35.91 & 6.56 \\
& Query Prediction Fisher Information
& 42.64 & 53.79
& 45.51 & 50.49
& 34.31 & 37.97
& 39.09 & 40.49
& 70.02 & 69.41
& \cellcolor{hlpink}\bestval{32.97} & 35.48
& 23.88 & 24.05
& 21.91 & 24.66
& 40.42 & 4.81 \\
& Support Prediction mIoU
& 39.96 & 57.29
& 45.93 & 57.67
& 43.81 & 61.22
& \cellcolor{hlpink}\bestval{74.94} & \cellcolor{hlpink}\bestval{75.75}
& 73.38 & 84.31
& 31.64 & \cellcolor{hlpink}\bestval{37.90}
& 43.48 & \cellcolor{hlpink}\bestval{55.60}
& 33.90 & 57.25
& 54.63 & 2.50 \\
& Reverse mIoU Consistency
& 41.28 & 55.00
& 46.61 & 55.05
& 46.63 & 58.30
& 74.08 & 74.78
& 76.17 & 83.11
& 31.20 & 36.26
& 41.24 & 53.45
& 38.97 & 57.06
& 54.32 & 2.81 \\
& Gram Matrix Distance
& 38.24 & 44.72
& 39.33 & 43.61
& 30.36 & 36.38
& 37.00 & 41.43
& 49.84 & 54.21
& 25.14 & 30.50
& 22.36 & 24.07
& 16.25 & 18.60
& 34.50 & 6.50 \\
& OT F1 Score
& 46.58 & 55.73
& 46.53 & 50.85
& 43.01 & 50.33
& 57.38 & 56.87
& 79.03 & 82.33
& 31.98 & 34.52
& 37.46 & 34.89
& 36.60 & 44.84
& 49.31 & 3.50 \\
& OT-BLS (ours)
& \bestval{47.70} & \bestval{58.37}
& \bestval{51.30} & \cellcolor{hlpink}\bestval{62.51}
& \cellcolor{hlpink}\bestval{52.13} & \cellcolor{hlpink}\bestval{62.62}
& 72.93 & 73.91
& \cellcolor{hlpink}\bestval{84.05} & \cellcolor{hlpink}\bestval{86.04}
& 31.69 & 34.96
& \cellcolor{hlpink}\bestval{44.26} & 52.17
& \cellcolor{hlpink}\bestval{46.92} & \cellcolor{hlpink}\bestval{60.33}
& \cellcolor{hlpink}\bestval{57.62} & \bestval{1.69} \\
\hline
\end{tabular}
}
\end{table*}

\subsection{Selection Criterion}
We now define the OT-BLS score $s^{(l)}$ used in Eq.~\eqref{eq:selected_layer}.
For a candidate layer $l\in\mathcal{L}$, the score of the candidate mask $\hat{m}_q^{(l)}$ is
\begin{equation}
s^{(l)} = \alpha\,\precision_{\fg}^{(l)} + (1-\alpha)\,\precision_{\bg}^{(l)},
\label{eq:score}
\end{equation}
where $\alpha\in[0,1]$ controls the balance between foreground and background label agreement.
Both terms are precision scores.
Let $\mathrm{TP}_{\fg}^{(l)}$ be the amount of the predicted query foreground matched to the support foreground, and let $\mathrm{FP}_{\fg}^{(l)}$ be the amount matched to the support background.
We define $\mathrm{TP}_{\bg}^{(l)}$ and $\mathrm{FP}_{\bg}^{(l)}$ analogously for the background.
The two precisions are then
\begin{align}
\precision_{\fg}^{(l)}
&= \frac{\mathrm{TP}_{\fg}^{(l)}}{\mathrm{TP}_{\fg}^{(l)} + \mathrm{FP}_{\fg}^{(l)}},
\label{eq:precision_fg} \\
\precision_{\bg}^{(l)}
&= \frac{\mathrm{TP}_{\bg}^{(l)}}{\mathrm{TP}_{\bg}^{(l)} + \mathrm{FP}_{\bg}^{(l)}}.
\label{eq:precision_bg}
\end{align}
The score $s^{(l)}$ is a support--query label-agreement score.
It is high when the predicted query foreground matches the support foreground, and the predicted query background matches the support background.
Specifically, $\precision_{\fg}^{(l)}$ measures the fraction of the predicted query foreground matched to the support foreground rather than the support background, and $\precision_{\bg}^{(l)}$ measures the analogous fraction for the background.
Both terms are evaluated from the query side against the support labels, because no query ground truth is available.
A natural alternative is to combine each precision with its corresponding recall to compute an F1 score.
We avoid this design because foreground recall would be normalized by the amount of support foreground rather than by the query ground truth.
Such an F1 score would favor masks whose foreground area matches that of the support, which is inappropriate for CD-FSS, where target size can differ greatly between support and query images.

\subsection{Computing the Selection Criterion}
Computing $\mathrm{TP}_{\fg}^{(l)}$ and $\mathrm{FP}_{\fg}^{(l)}$ requires correspondences between query and support tokens.
We regard a query token and a support token as more likely to correspond when their features are more similar.
OT establishes a soft set-level correspondence from such similarities rather than relying on independent nearest-neighbor matches, and can therefore compare feature sets with different numbers of tokens and region sizes.
We therefore adopt OT and use the resulting plan as a patch-level correspondence between the query and the concatenated collection of support tokens.
For two feature sets $A=\{\mathbf{f}_i\}_{i=1}^{m}$ and $B=\{\mathbf{g}_j\}_{j=1}^{n}$ with cost matrix $C\in\mathbb{R}_{+}^{m\times n}$, we compute the entropy-regularized transport plan~\cite{cuturi2013sinkhorn}
\begin{equation}
P^* = \mathop{\arg\min}_{P\in U(a,b)} \sum_{i=1}^{m}\sum_{j=1}^{n} P_{ij}C_{ij} - \varepsilon\,H(P),
\label{eq:ot_plan}
\end{equation}
where
\begin{equation}
H(P)=-\sum_{i=1}^{m}\sum_{j=1}^{n} P_{ij}\bigl(\log P_{ij}-1\bigr)
\end{equation}
is the entropy term.
Here, $P_{ij}$ denotes the amount of mass transported from $\mathbf{f}_i$ to $\mathbf{g}_j$, $\varepsilon>0$ is the regularization coefficient, and $a$ and $b$ are the marginal weights.
The feasible set is
\begin{equation}
U(a,b)=\{P\in\mathbb{R}_{+}^{m\times n}\mid P\mathbf{1}_n=a,\; P^{\top}\mathbf{1}_m=b\}.
\end{equation}
In OT-BLS, we use the entries of $P^*$ rather than the objective value itself.

We compute these precisions from a single transport plan obtained before applying any foreground--background split.
The plan is computed from the last-layer patch-token features of the frozen DINOv3 backbone, which serve only to define this common patch-level correspondence.
If the feature space used for OT varied across candidate layers, score differences would reflect layer-specific representations rather than the candidate masks themselves.
The masks $\hat{m}_q^{(l)}$ are still produced from their respective layers.
Let $F_q^{(L)}=\{\mathbf{f}_i\}_{i=1}^{n_q}$ be the query feature tokens.
Let $F_{s,k}^{(L)}=\{\mathbf{g}_{k,j}\}_{j=1}^{n_k}$ be the support feature tokens of the $k$-th support image.
We concatenate all support tokens as
\begingroup
\setlength{\abovedisplayskip}{3pt}
\setlength{\belowdisplayskip}{3pt}
\setlength{\abovedisplayshortskip}{2pt}
\setlength{\belowdisplayshortskip}{2pt}
\begin{equation}
F_s^{(L)} = [F_{s,1}^{(L)}; F_{s,2}^{(L)}; \cdots; F_{s,K}^{(L)}]
= \{\mathbf{g}_j\}_{j=1}^{n_s},
\quad n_s=\sum_{k=1}^{K} n_k .
\label{eq:concat_support}
\end{equation}
\endgroup
Concatenating support tokens avoids constructing a separate correspondence for each support image.
Because different support examples may contain different query-relevant patches, the concatenated collection allows each query token to send mass to the most compatible support tokens across all examples.

For tokens in $F_q^{(L)}$ and $F_s^{(L)}$, we define the cost matrix using cosine distance as
\begingroup
\setlength{\abovedisplayskip}{3pt}
\setlength{\belowdisplayskip}{3pt}
\setlength{\abovedisplayshortskip}{2pt}
\setlength{\belowdisplayshortskip}{2pt}
\begin{equation}
C_{ij}=1-\cos(\mathbf{f}_i,\mathbf{g}_j).
\label{eq:cost}
\end{equation}
\endgroup
We assign uniform weights $a_i=1/n_q$ and $b_j=1/n_s$ to all tokens and solve Eq.~\eqref{eq:ot_plan}.
The resulting transport plan $P^*$ gives the transported mass $P^*_{ij}$ from query token $\mathbf{f}_i$ to concatenated support token $\mathbf{g}_j$.

\begin{table}[!t]
\centering
\scriptsize
\setlength{\tabcolsep}{3pt}
\caption{Average runtime per episode (ms).
The fastest value in each shot setting is bold.
Rank is computed by sorting runtimes in ascending order among the selection strategies.}
\label{tab:runtime}
\begin{tabular}{lccc}
\hline
Selection strategy & 1-shot & 5-shot & Rank \\
\hline
Entropy & 2.0 & 2.0 & 2.00 \\
Regularized Patch-Norm Ratio & \bestval{0.2} & \bestval{0.4} & \bestval{1.00} \\
Query Prediction Fisher Information & 10.4 & 10.5 & 4.00 \\
Support Prediction mIoU & 17.6 & 75.9 & 6.00 \\
Reverse mIoU Consistency & 293.7 & 361.9 & 7.00 \\
Gram Matrix Distance & 14.3 & 15.7 & 5.00 \\
OT-BLS (ours) & 4.8 & 5.2 & 3.00 \\
\hline
\end{tabular}
\vspace{-1.7em}
\end{table}

Foreground and background labels are used only after $P^*$ has been obtained.
For a candidate layer $l$, the patch-grid version of $\hat{m}_q^{(l)}$ assigns foreground and background labels to query-token positions for score aggregation only.
We denote the query positions labeled as foreground and background by $\mathcal{I}_{q,\fg}^{(l)}\subseteq\{1,\dots,n_q\}$ and $\mathcal{I}_{q,\bg}^{(l)}\subseteq\{1,\dots,n_q\}$.
The support annotations, downsampled to the same patch grid, define $\mathcal{I}_{s,\fg}\subseteq\{1,\dots,n_s\}$ and $\mathcal{I}_{s,\bg}\subseteq\{1,\dots,n_s\}$ for the concatenated support tokens.
For two index sets $A\subseteq\{1,\dots,n_q\}$ and $B\subseteq\{1,\dots,n_s\}$, we define
\begin{equation}
P^*(A,B)=\sum_{i\in A}\sum_{j\in B}P^*_{ij},
\label{eq:region_mass}
\end{equation}
which is the total transported mass from query-token indices in $A$ to support-token indices in $B$.
Although $P^*$ is shared, the query-token labels used for aggregation change with $\hat{m}_q^{(l)}$.
Thus, each candidate mask sums different rows of the same transport plan.
Under this plan, the true-positive and false-positive amounts in Eqs.~\eqref{eq:precision_fg} and~\eqref{eq:precision_bg} become transported masses.
Specifically, $\mathrm{TP}_{\fg}^{(l)}=P^*(\mathcal{I}_{q,\fg}^{(l)}, \mathcal{I}_{s,\fg})$ and $\mathrm{FP}_{\fg}^{(l)}=P^*(\mathcal{I}_{q,\fg}^{(l)}, \mathcal{I}_{s,\bg})$, with analogous definitions for the background.
Substituting these masses into Eqs.~\eqref{eq:precision_fg} and~\eqref{eq:precision_bg},
with a small constant $\delta$ added to each denominator, gives
\begingroup
\setlength{\abovedisplayskip}{3pt}
\setlength{\belowdisplayskip}{3pt}
\setlength{\abovedisplayshortskip}{2pt}
\setlength{\belowdisplayshortskip}{2pt}
\medmuskip=2mu
\begin{align}
\precision_{\fg}^{(l)}
&=
\frac{P^*(\mathcal{I}_{q,\fg}^{(l)}, \mathcal{I}_{s,\fg})}
{P^*(\mathcal{I}_{q,\fg}^{(l)}, \mathcal{I}_{s,\fg})
+ P^*(\mathcal{I}_{q,\fg}^{(l)}, \mathcal{I}_{s,\bg}) + \delta},
\label{eq:precision_fg_ot} \\
\precision_{\bg}^{(l)}
&=
\frac{P^*(\mathcal{I}_{q,\bg}^{(l)}, \mathcal{I}_{s,\bg})}
{P^*(\mathcal{I}_{q,\bg}^{(l)}, \mathcal{I}_{s,\bg})
+ P^*(\mathcal{I}_{q,\bg}^{(l)}, \mathcal{I}_{s,\fg}) + \delta},
\label{eq:precision_bg_ot}
\end{align}
\endgroup
where $\delta=10^{-12}$ prevents division by zero when a predicted region is empty.

\begin{figure}[!t]
\centering
\includegraphics[width=\linewidth]{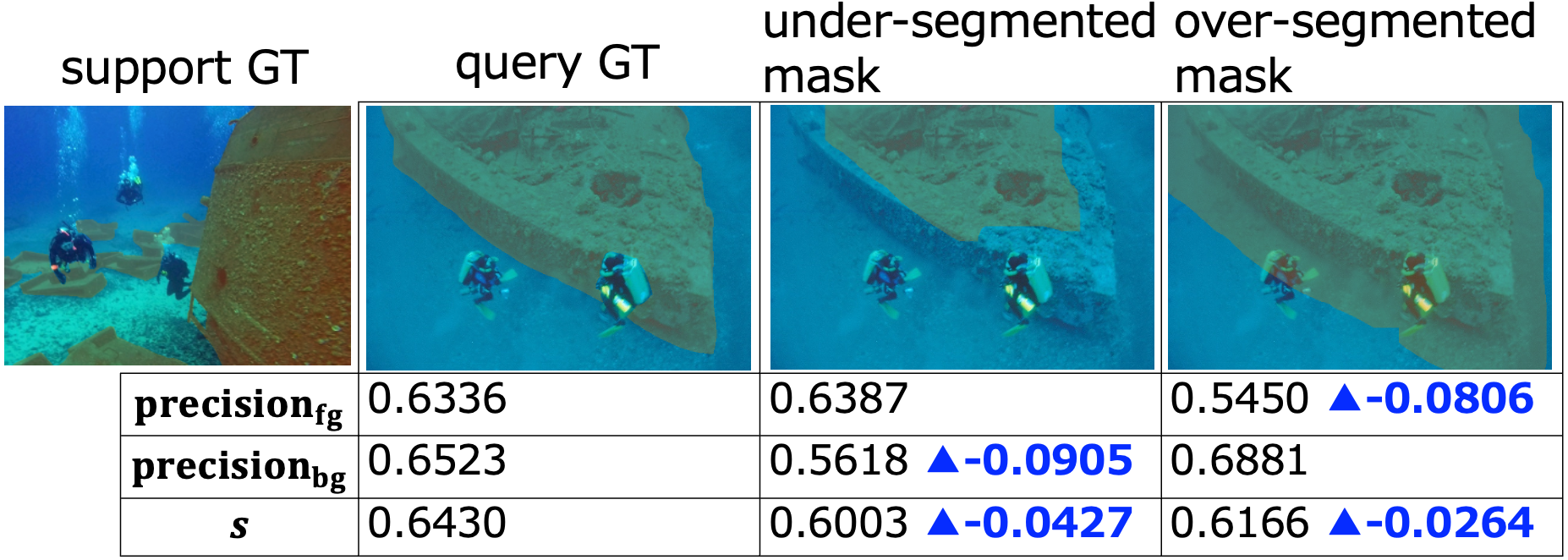}
\caption{Values of $\precision_{\fg}$, $\precision_{\bg}$, and $s$ for ground-truth, under-segmented, and over-segmented masks.}
\label{fig:segmentation_cases}
\vspace{-0.7em}
\end{figure}

The denominators remove query-side size bias.
Under uniform query weights, a predicted region with more tokens emits more mass.
Dividing by the emitted mass makes the foreground and background precisions comparable.
Thus, each precision is the fraction of mass emitted from a predicted query region that is transported to the support region with the same label.
Maximizing $s^{(l)}$ therefore selects the layer whose foreground--background partition best agrees with the support labels under $P^*$.
Finally, we restrict the search set to $\mathcal{L}=\{\lfloor L/2\rfloor+1,\dots,L\}$ because prior work reports that objectness and semantic information are concentrated in later layers, which benefits support--query consistency in CD-FSS~\cite{ma2025hera}.

\begin{figure}[!t]
\centering
\includegraphics[width=\linewidth]{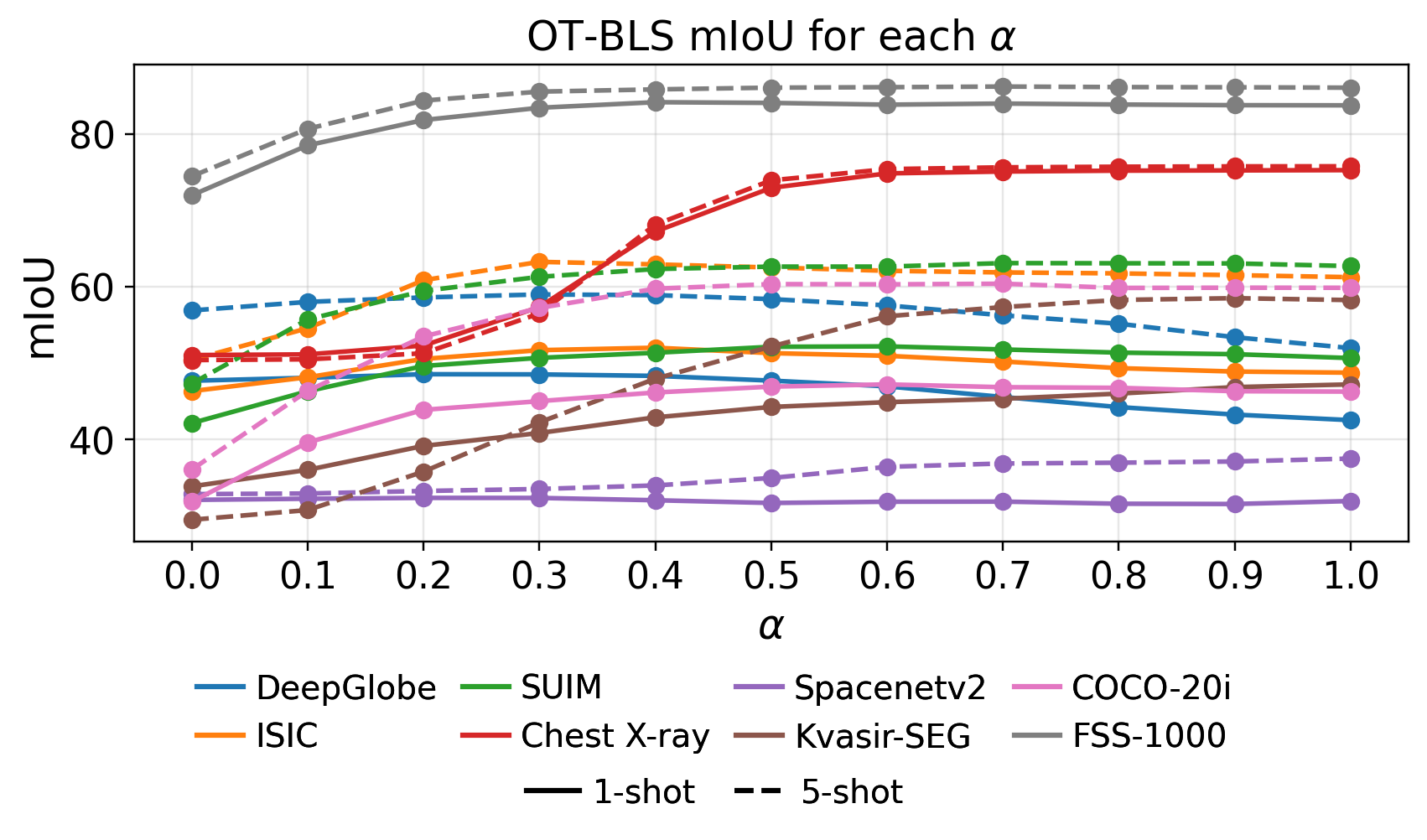}
\caption{Effect of the foreground--background weighting parameter $\alpha$ in Eq.~\eqref{eq:score} on OT-BLS mIoU for 1-shot and 5-shot evaluation across the eight benchmarks.}
\label{fig:precision_recall}
\end{figure}

\section{Experiments}
\subsection{Experimental Setup}
We follow the experimental setting of Zakir \textit{et al.}~\cite{zakir2026revealingsemanticselectiongap}.
The backbone is a frozen DINOv3 ViT-B with an input resolution of 512 $\times$ 512.
For ViT-B with $L=12$, the search set is $\mathcal{L}=\{7,\dots,12\}$.
Mean intersection over union (mIoU) is the evaluation metric for CD-FSS.
Because each episode contains a single target class, mIoU here is not an average over multiple classes within an episode.
For each method, we aggregate confusion matrices over all episodes, exclude the background class, and compute mIoU from the aggregated counts rather than averaging episode-wise mIoU scores.
We use eight benchmarks covering standard and cross-domain FSS settings.
COCO-20i~\cite{lin2014coco} serves as the standard FSS benchmark and contains 80 natural-image object classes.
The other datasets are DeepGlobe~\cite{demir2018deepglobe} for satellite imagery, ISIC~\cite{codella2019isic} for skin lesions, SUIM~\cite{islam2020suim} for underwater scenes, FSS-1000~\cite{li2020fss1000}, Chest X-ray~\cite{indeewara2023chest}, SpaceNetv2~\cite{etten2018spacenet}, and Kvasir-SEG~\cite{jha2020kvasir} for gastrointestinal polyps.
For COCO-20i, we split the 80 classes into four folds, randomly sample 1,000 episodes per fold, and report mIoU averaged over four folds.
Here, $K$ is the number of support image--mask pairs in each episode, and a setting using $K$ pairs is called $K$-shot.
For other datasets, we set $K\in\{1,5\}$ and randomly sample 600 episodes for each $K$-shot setting.

In Eq.~\eqref{eq:score}, we set $\alpha=0.5$ unless otherwise stated, using a neutral default that gives equal weight to foreground and background agreement.
Figure~\ref{fig:precision_recall} analyzes the effect of varying $\alpha$.
We run the Sinkhorn algorithm with $\varepsilon=0.05$, a maximum of 200 iterations, and a marginal-error tolerance of $10^{-6}$.
Runtime is measured on one NVIDIA H200 GPU after the shared DINOv3 forward pass and candidate-mask generation.

\subsection{Comparison Methods}
To assess layer-selection performance within a fixed prediction pipeline, we compare OT-BLS with six criteria from Zakir \textit{et al.}~\cite{zakir2026revealingsemanticselectiongap}, all of which select from the same candidate masks in each episode.
These are Entropy, Regularized Patch-Norm Ratio, Query Prediction Fisher Information, Support Prediction mIoU, Reverse mIoU Consistency, and Gram Matrix Distance.
We report the last-layer baseline, a query-ground-truth-guided oracle upper bound, and an ablation of the scoring design.
OT F1 Score replaces each precision in Eq.~\eqref{eq:score} with an F1 score that adds the corresponding support-normalized recall, thereby testing the effect of incorporating recall into the score.

\subsection{Results and Discussion}
Table~\ref{tab:main} reports the 1-shot and 5-shot results.
OT-BLS outperforms the last-layer baseline in 12 of 16 dataset/shot settings.
It falls below the baseline only on DeepGlobe (1- and 5-shot), ISIC 1-shot, and Kvasir-SEG 5-shot.
Compared with existing criteria, OT-BLS achieves the best mIoU in 11 of 16 settings and the best overall average rank.
These results suggest that transport-based support--query label agreement is more reliable than proxy-based layer selection in most settings.

Table~\ref{tab:runtime} shows that OT-BLS ranks third in runtime and is faster than Support Prediction mIoU, the best-performing existing criterion, by 3.7$\times$ in 1-shot and 14.6$\times$ in 5-shot.
Together with Table~\ref{tab:main}, this result indicates that OT-BLS provides the best selection performance with a modest runtime cost.

Figure~\ref{fig:segmentation_cases} compares $\precision_{\fg}$, $\precision_{\bg}$, and $s$ for ground-truth, under-segmented, and over-segmented masks.
The under-segmented mask decreases $\precision_{\bg}$, whereas the over-segmented mask decreases $\precision_{\fg}$.
This indicates that $\precision_{\bg}$ captures query-background agreement and $\precision_{\fg}$ captures query-foreground consistency with the support labels.
The final score $s$ is also lower for both erroneous masks.

Figure~\ref{fig:precision_recall} shows the mIoU of OT-BLS as we sweep the weighting parameter $\alpha$ in Eq.~\eqref{eq:score} from $0.0$ to $1.0$ in increments of $0.1$.
The curves compare the resulting layer-selection performance for both 1-shot and 5-shot settings across all benchmarks.
The performance changes noticeably with $\alpha$, and the best $\alpha$ differs across datasets and shot settings.
This suggests that the appropriate balance between foreground and background agreement depends on the target domain.
In the main experiments, we therefore use $\alpha=0.5$ as a neutral default and leave adaptive selection of $\alpha$ for future work.

\section{Conclusion}
We have proposed OT-BLS, an optimal-transport-based layer-selection criterion for CD-FSS with frozen DINOv3.
OT-BLS computes a support--query transport plan and scores each candidate mask by aggregating transported mass across foreground and background regions.
The score evaluates whether the predicted query labels agree with the support labels under patch-level correspondence.
Experiments on eight benchmarks show that OT-BLS outperforms the last-layer baseline in 12 of 16 dataset/shot settings, improves the average mIoU from 55.72 to 57.62, and achieves the best average rank among the evaluated criteria.
These results indicate that transport-based support--query label agreement is a more reliable layer-selection signal than the proxy-based criteria used in prior work.

\end{document}